\documentclass[journal]{IEEEtran}
\ifCLASSINFOpdf
\else
\fi
\usepackage{microtype}
\usepackage{graphicx}
\usepackage{subfigure}
\usepackage{booktabs}
\usepackage{amsmath}

\begin{document}
%
\title{Self-supervised learning for autonomous vehicles perception: A conciliation between analytical and learning methods}
%
%
%

\author{Florent~Chiaroni$^{\star \ddagger}$,
		Mohamed-Cherif~Rahal$^{\star}$,
		Nicolas~Hueber$^{\dagger}$,
        and Fr\'ed\'eric~Dufaux$^{\ddagger}$
        
\thanks{$^{\star}$VEDECOM Institute, Department of delegated driving (VEH08), Perception team, \{florent.chiaroni, mohamed.rahal\}@vedecom.fr}
\thanks{$^{\dagger}$French-German Research Institute of Saint-Louis (ISL), ELSI team, nicolas.hueber@isl.eu}
\thanks{$^{\ddagger}$Universit\'e Paris-Saclay, CNRS, CentraleSupelec, Laboratoire des signaux et syst\`emes, \{florent.chiaroni, frederic.dufaux\}@centralesupelec.fr}}

%
%

\markboth{Accepted - IEEE Signal Processing Magazine - Special Issue on Autonomous Driving}%
{Shell \MakeLowercase{\textit{et al.}}: Bare Demo of IEEEtran.cls for IEEE Journals}
%



\maketitle



%
\IEEEpeerreviewmaketitle

\section{Introduction}
%
%
%
%

%

%
\IEEEPARstart{T}{he} interest for autonomous driving has continuously increased during the last two decades. However, to be adopted, such critical systems need to be safe. Concerning the perception of the ego-vehicle environment, the literature has investigated two different types of methods. On the one hand, traditional analytical methods generally rely on hand-crafted designs and features. On the other hand, learning methods aim at designing their own appropriate representation of the observed scene.

\textbf{Analytical methods} have demonstrated their usefulness for several tasks, including keypoints detection \cite{lowe_distinctive_2004}, \cite{karpushin_keypoint_2016}, optical flow, depth map estimation, background subtraction, geometric shape detection, tracking, and simultaneous localization and mapping (SLAM) \cite{bresson_simultaneous_2017}. Those methods have the advantage to be easily explainable. However, it is difficult to apply them on high dimensional data for semantic scene analysis. For example, identifying the other road users or understanding the large variety of situations present in an urban scene requires to extract complex patterns from high dimensional data captured by camera sensors.

\textbf{Learning methods} are nowadays the most adapted in terms of prediction performances for complex pattern recognition tasks \cite{kirillov_panoptic_2018} implied in autonomous vehicles scene analysis and understanding. However, state-of-the-art results are often obtained with large and fully labeled training datasets \cite{cordts_cityscapes_2016}. Hand-labeling a large dataset for a given specific application has a cost. Another difficulty is to apprehend from end-to-end the learned representations. To overcome the former limitation, transfer learning and weakly supervised learning methods have been proposed. Some of them can exploit partially labeled datasets \cite{niu_theoretical_2016}, \cite{chiaroni_learning_2018}, or noisy labeled datasets \cite{ma_dimensionality-driven_2018}, \cite{chiaroni_hallucinating_2019}. Concerning the latter problem, under mild theoretical assumptions on the learning model, we can interpret the predicted outputs. For instance, it is possible to automatically detect overfitting of the training \cite{houle_local_2017}, to estimate the fraction of mislabeled examples \cite{jain_estimating_2016}, or to estimate the uncertainty in the prediction outputs \cite{gal_uncertainty_2016}.

In addition to the difficulty of obtaining a large labeled training dataset, another challenge of learning methods is to \textbf{prevent unpredictable events}. Indeed, some scenes unseen during the training can appear frequently in the context of the autonomous vehicle. For instance, an accident on the road can change drastically the appearance and the location of potential obstacles. Thus, even if it is possible to predict when the model does not know what it observes, it may be interesting to confirm it through an analytical process and to adapt the learning model to this novel situation.

It turns out that \textbf{self-supervised learning methods (SSL)}, consisting of combining analytical and learning techniques, have shown in the literature the ability to address such issues. For instance, the SSL system in \cite{dahlkamp_self_supervised_2006} won the 2005 DARPA Grand Challenge thanks to its adaptability to changing environments. SSL for autonomous driving vehicles perception is most often based on learning from data which is automatically labeled by an upstream method, similarly to feature learning in \cite{jing_self-supervised_2019}. In this paper, we address the following aspects of SSL:
\begin{itemize}
	\item abilities such as 
	sequential environment adaptation on the application time, referred to as online learning, self-supervised evaluation, non-necessity of hand-labeled data, fostering of multimodal techniques \cite{dahlkamp_self_supervised_2006}, and self-improvement. For example, iterative learning reduces progressively the corrupted predictions \cite{zhong_self_supervised_2017};
	\item tasks made possible thanks to those advantages, such as depth map estimation \cite{garg_unsupervised_2016}, \cite{zhong_self_supervised_2017}, temporal predictions \cite{dequaire_deep_2017}, moving obstacles analysis \cite{bewley_online_2014}, and long-range vision \cite{dahlkamp_self_supervised_2006}, \cite{hadsell_learning_2009}. For example, the SSL system in \cite{hadsell_learning_2009} learns to extrapolate the appearance of obstacles and traversable areas observable by stereo-vision in a short-range, to identify them at a longer distance beyond the detection range of the stereo-vision.
\end{itemize} 

\begin{figure*}[h]
\begin{center}
\begin{minipage}[c]{0.321\linewidth}
  \centering
  \centerline{\includegraphics[width=\linewidth]{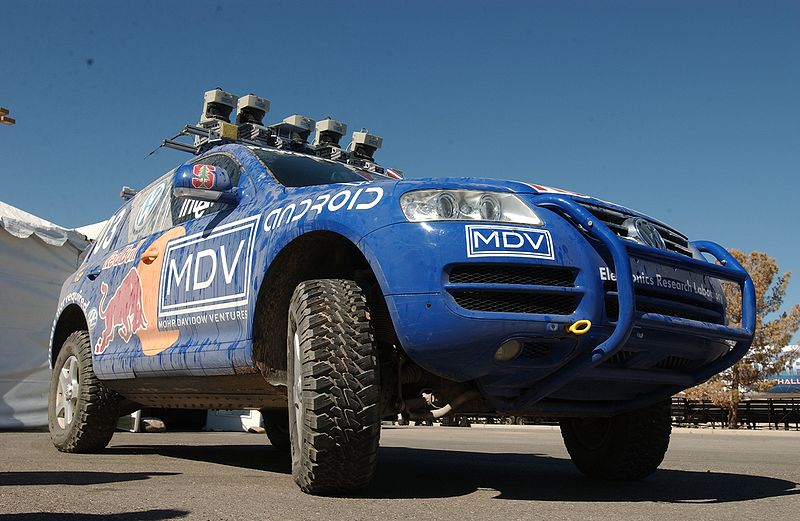}}
  \centerline{(a)}\medskip
\end{minipage}
\begin{minipage}[c]{0.25\linewidth}
  \centering
  \centerline{\includegraphics[width=\linewidth]{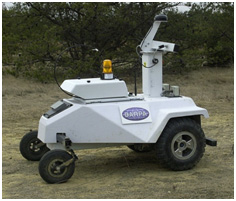}}
  \centerline{(b)}\medskip
\end{minipage}
\begin{minipage}[c]{0.215\linewidth}
  \centering
  \centerline{\includegraphics[width=\linewidth]{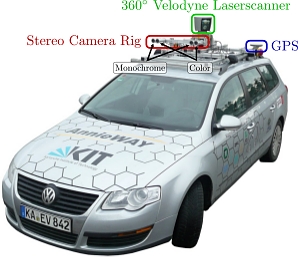}}
  \centerline{(c)}\medskip
\end{minipage}
\caption{Some self-driving cars. (a) is the self-driving car \textit{Stanley} that won the \textit{DARPA Grand Challenge} using a SSL system equipped with a calibrated monocular camera and a LIDAR sensor \cite{dahlkamp_self_supervised_2006}. (b) is the autonomous mobile robot \textit{LAGR}. It integrates another SSL vision approach \cite{hadsell_learning_2009} able to identify online the obstacles and road segmentation from a short-range stereovision up to a long-range monocular vision. (c) is the car equipped with the perception sensors used to generate the KITTI dataset \cite{geiger_are_2012}.}
\label{fig:Self_driving_cars}
\end{center}
%
\end{figure*}

While the cited SSL techniques are respectively designed for a specific use case application, they present some similarities. In particular, a shared underlying idea is to: Learn to predict, from a given spatio-temporal information (e.g. a single camera frame \cite{dahlkamp_self_supervised_2006}, \cite{hadsell_learning_2009}, \cite{guizilini_online_2013}, \cite{garg_unsupervised_2016}, \cite{pathak_learning_2017}), something (e.g. traversable area segmentation \cite{dahlkamp_self_supervised_2006}, \cite{hadsell_learning_2009}, depth estimation \cite{garg_unsupervised_2016}, or moving obstacles segmentation \cite{guizilini_online_2013}, \cite{pathak_learning_2017}) that can be automatically labeled in another way using additional spatio-temporal information (e.g. stereo-vision camera \cite{hadsell_learning_2009}, \cite{garg_unsupervised_2016}, a temporal sequence \cite{ondruska_deep_2016}, or depth sensor \cite{dahlkamp_self_supervised_2006}).

We propose to highlight those inter-dependencies hereafter. In this way, we aim at providing to the reader some analytical, learning and hybrid tools which are transversal to the final application use cases. In addition, the limitations of the presented frameworks are discussed and highlighted in Table \ref{tab:differences_A_L_SSL}, as well as the perspectives of improvement for self-evaluation, self-improvement, and self-adaptation, in order to address future autonomous driving challenges.

\begin{table*}[h]
\centering
\caption{Comparison of state-of-the-art Analytical, Learning and SSL methods for autonomous vehicle perception challenges ('+' inappropriate, '++' intermediary, '+++' appropriate).}
\resizebox{17.2cm}{!}{
\begin{tabular}{c||c|c|c|c|c}
\toprule
Methodology & no hand-labeling  & dense complex pattern analysis  & online self-evaluation and adaptation  & knowledge extrapolation & low-cost sensor requirements \\
\midrule
\midrule
Analytical & +++ & + & ++ & + & +\\
Supervised learning & + & +++ & + & + & +++\\
Self-Supervised Learning & +++ & ++ & +++ & +++ & ++ \\
\bottomrule
\end{tabular}}
\label{tab:differences_A_L_SSL}
\end{table*}

The outline of this article is as follows. After this introduction, we present in Sec. \ref{section:analytical} and \ref{section:learning} some analytical and learning perception tools relevant to SSL. We follow in Sec. \ref{section:ssl} with the presentation of existing SSL techniques for some autonomous driving perception applications. Finally, we conclude with a discussion focusing on limitations and future challenges in Sec. \ref{section:limit}.

\section{Analytical methods}
\label{section:analytical}

Before the recent growing interest for deep learning methods, many analytical methods (without learning) have been proposed, bringing baseline reference tools for multiple challenging perception tasks in the context of autonomous driving. Some of the most investigated tasks considered in this article are briefly introduced hereafter:

\begin{itemize}
	\item \textbf{Keypoints feature detection:} Before analyzing the sensor data from a relatively high level, analytical techniques often require to perform spatial or temporal data matching using \textbf{feature detection} methods. More specifically, these methods consist in detecting and extracting local features in the sensor data. These hand-crafted features can be small regions of interest \cite{harris1988combined}. In order to enable the matching of sensor data, captured from the same scene with different spatial or temporal points of view, such features need to be as invariant as possible to scale, translation, and rotation transformations. 
	The most common sensor data is an image captured by a camera. In this case, competitive feature detectors include SIFT \cite{lowe_distinctive_2004}, SURF \cite{bay2006surf}, and ORB \cite{rublee2011orb}. When a depth sensor is also available, the depth information can be exploited in order to further improve feature detection. For instance, the TRISK method \cite{karpushin_keypoint_2016} is specifically designed for RGB-D images. More recently, LIDAR has enabled the acquisition of point clouds. To tackle this new form of sensor data, some feature detection techniques are derived from image ones (e.g. Harris and SIFT). Alternatively, some new approaches such as ISS \cite{zhong2009intrinsic} are exclusively designed for point clouds.
	From a practical point of view, implementations of common image feature detectors can be found in image libraries as OpenCV\footnote{https://opencv.org/}, and in point clouds libraries as PCL\footnote{http://pointclouds.org/}. Feature detectors are exploited by several autonomous driving perception techniques requiring matching of sensor data, including optical flow, disparity map, visual odometry, SLAM and tracking techniques.
	\item \textbf{Optical flow} is a dense \cite{farneback_two-frame_2003} or sparse \cite{lucas1981iterative} motion pattern. It can be obtained by computing points or features transformations throughout a temporal images sequence captured from a static or mobile ego-camera point of view. In the context of autonomous driving perception, optical flow is relevant for background subtraction, motion estimation of the ego-vehicle and surrounding moving obstacles as proposed by Menze et al. \cite{menze2015object}. It can also be exploited, in the case of a monocular mobile camera without any additional information, for relative depth map estimation \cite{prazdny1980egomotion} of the surrounding static environment.
	\item \textbf{Depth map estimation} aims at providing image pixels depths, namely the relative or absolute distance between the camera and the captured objects. Several techniques exist to address this task. One of the most common and effective approaches is to compute a disparity map from a stereo-camera.
	Combined with the extrinsic cameras parameters, such as the baseline distance separating both cameras, the disparity map can be converted into an inversely proportional absolute depth map. Another approach is to project LIDAR points  on some of the camera image pixels. It also requires extrinsic spatial and temporal calibrations between both sensors. As mentioned previously, a relative depth map of a static scene can also be derived from the optical flow obtained with a moving camera. Under some assumptions, for example with additional accurate GPS and IMU sensors information concerning the absolute pose transformations of the moving camera, the absolute depth map can then be obtained. The depth map can also be directly obtained with some RGB-D sensors. Depth map is interesting for identifying the 3D shape of objects in the scene. More specifically, in autonomous driving, an absolute depth map is relevant for estimating the distance between the ego-vehicle and detected obstacles. However, we should note that absolute depth map estimation is more challenging compared to relative depth map, as at least two jointly calibrated sensors are necessary. Consequently, it implies a relative higher financial cost in production. Moreover, extrinsic calibrations can be sensitive to the ego-vehicle physical shocks. Finally such sensor fusions can only offer depth estimation in a limited range, due to fixed baselines with stereo cameras, or sparse point cloud projections with dispersive LIDAR sensors. Nevertheless, relative depth map is sometimes sufficient to detect obstacles and traversable areas. For example, considering the traversable area as a set of planes in the depth map 3D point cloud projection, some template matching techniques can be used \cite{hadsell_learning_2009}.
	\item \textbf{Geometric shape detection} techniques such as Hough transform and RANSAC \cite{fischler1981random} initially aimed at identifying some basic geometric shapes such as lines for lane marking detection, ellipses for traffic lights detection, or planes for road segmentation. In order to deal with sophisticated template matching tasks, techniques such as the Hough transform have been generalized for arbitrary shape detection \cite{ballard1981generalizing}. Nonetheless, these techniques require an exact model definition of the shapes to be detected. Consequently, they are sensitive to noisy data and are impractical for detection of complex and varying shapes such as obstacles encountered in the context of autonomous driving. Indeed, such objects typically suffer from outdoor illumination changes, background clutter, or non-rigid transformations.
	\item \textbf{Motion tracking} aims at following some data points, features or objects through time. Tracking filters, such as the Extended Kalman Filter (EKF), predict the next motion using the prior motion knowledge.
	Conversely, objects tracking can be achieved by features or template matching between consecutive video frames. Pixel points and features tracking is relevant for dense or sparse optical flow, as well as visual odometry estimation \cite{scaramuzza2011visual}. Conversely, obstacle objects tracking is very important in autonomous driving for modeling or anticipating their trajectories into the ego-vehicle environment. However, on the whole, while some techniques integrate uncertainty, they remain limited when dealing with complex real motion patterns. Pedestrians and drivers behaviour prediction typically requires knowledge about the context. Moreover, mobile obstacles appearance can drastically change depending on their orientation.
	\item \textbf{SLAM techniques:} The complementarity between the above enumerated concepts has been demonstrated through the problem of \textit{simultaneously} \textit{localizing} the ego-vehicle \textit{and mapping} the surrounding environment (SLAM) \cite{bresson_simultaneous_2017}. Features matching provides the pose transformations of the moving ego-vehicle. In turn, 3D scaled projections of depth maps combined with the successive estimated poses provide the environment mapping. Tracking filters and template matching may offer some robustness against sensor data noise and drifting localization estimation, as respectively proposed in EKF SLAM \cite{davison2007monoslam} and SLAM$++$ \cite{Salas_Moreno_2013_CVPR} approaches.
\end{itemize}
To summarize, analytical methods can successfully deal with several perception tasks of significant interest in the context of autonomous driving. In particular, a self-driving vehicle embedding these techniques is able to carry out physical analysis such as the 3D reconstruction modelling of the environment, and dynamic estimations concerning the ego-vehicle and the encountered surrounding mobile obstacles. Moreover, these techniques have the advantage to be easily explainable in terms of design. This facilitates the identification and prevention of failure modes. However, some critical limitations persist nowadays:
\begin{itemize}
	\item A lack of landmarks and salient features combined with the presence of dynamic obstacles may entail a severe degradation of the feature detection and matching.
	\item Severe noisy sensor data induces the same risks. 
	\item It is impossible to achieve dense real-time semantic scene analysis of environments including a wide range of complex shape patterns. 
\end{itemize}
Learning methods, by recognizing and predicting complex patterns with generalization abilities, aim at overcoming such issues, as developed in the next section. 

\section{Learning methods}
\label{section:learning}

Learning methods have demonstrated state-of-the-art prediction performances for semantic analysis tasks during the last decade. Autonomous driving is a key application which can greatly benefit from these recent developments. For instance, learning methods have been investigated in this context, for identifying the observed scene context using classification, for detecting the other road users surrounding the ego-vehicle, for delineating the traversable area surface, or for dynamic obstacles tracking.
\begin{itemize}
	\item \textbf{Classification} aims at predicting, for a given input sensor sample, an output class label. In order to deal with high dimensional data containing complex patterns, the first stage is generally to extract relevant features using hand-crafted filters or learned feature extractors. For image feature extraction, the state-of-the-art techniques use Convolutional Neural Network (CNN) architectures. They are composed of a superposition of consecutive layers of trainable convolutional filters. Then, a second stage is to apply a learning classifier on the feature maps generated as output of these filters. Some commonly used classifiers are the Support Vector Machine (SVM) and the Multi-Layer Perceptron (MLP). Both require a training which is most of the time performed in a fully supervised way on labeled data. The CNN and MLP deep learning models are trained by backpropagating the output prediction error on the trainable weigths up to the input. Concerning the evaluation of these models, a test dataset is required, which is labeled as well. The \textit{Accuracy} metric is commonly used for evaluating the prediction performances, while the F1-Score, an harmonic mean of the precision and recall, is relevant for information retrieval. An image classification application example in autonomous driving is for categorizing the context of the driven road \cite{teichmann_multinet:_2016}. 
	\item \textbf{Detection} generally localizes the regions of interest in a visual sensor data, which in turn can be classified. A commonly used strategy, invariant to scales and translations, applies an image classifier on sliding windows over an image pyramid. Then, several advanced competitive image detection techniques, such as Faster R-CNN \cite{ren_faster_2015} or Yolo \cite{redmon_you_2016} have been more recently developed, and have been adapted for road users detection \cite{teichmann_multinet:_2016}.
	\item \textbf{Segmentation:} As its name suggests, this task provides a segmentation of visual sensor data. Three distinct applications can be considered:  
	\begin{itemize}
		\item \textit{Semantic segmentation} assigns a semantic class label to each pixel. An example is road segmentation \cite{teichmann_multinet:_2016}. State-of-the-art methods for autonomous vehicle perception can exploit an auto-encoder architecture, but also dilated or atrous convolutions, as well as an image context modeling strategy as reviewed in \cite{garcia2018survey}.		
		In the context of image segmentation, these models are trained to predict as output a pixel-wise classification of the input image.
		\item \textit{Instance segmentation} aims at detecting and segmenting each object instance. Examples include foreground segmentation and object detection of potentially moving obstacles \cite{He_2017_ICCV}. 
		\item \textit{Panoptic segmentation} \cite{kirillov_panoptic_2018} is a unification of the two previously mentioned segmentation tasks.
	\end{itemize} 
	Some models dealing with these segmentation tasks have been adapted for performing per-pixel regression tasks such as dense optical flow estimation \cite{Dosovitskiy_2015_ICCV} or depth map estimation \cite{deep_sup_depth_map_est}.
	\item \textbf{Temporal object tracking} follows the spatial location of selected objects along a temporal data sequence. State-of-the-art learning techniques use variants of the Recurrent Neural Network (RNN) model \cite{milan_online_2017}. Compared to standard filtering techniques, RNNs have the ability to learn complex and relatively long-term temporal patterns in the context of autonomous driving.
\end{itemize} 
These methods can be combined in a unified framework, for instance by sharing the same encoded latent feature maps, as proposed in MultiNet \cite{teichmann_multinet:_2016} for joint real-time scene classification, vehicle detection and road segmentation.
While demonstrating competitive prediction performances, the above mentioned learning techniques are fully supervised. In other words, they have in common the limitation to require large-scale fully annotated training datasets. In order to alleviate this issue, some other learning strategies have been investigated:
\begin{itemize}
	\item \textbf{Weakly supervised learning:} These techniques can be trained with a partially labeled dataset \cite{niu_theoretical_2016}, and eventually with a fraction of corrupted labels \cite{ma_dimensionality-driven_2018}, \cite{chiaroni_hallucinating_2019}. Advantageously, these approaches drastically reduce the need of labeled data.
	\item \textbf{Clustering}: These approaches can be defined as an unlabeled classification strategy that aims at gathering without supervision the data depending on their similarities. A huge advantage is that no labels are required. However, if it is necessary to associate the resulting clusters with semantic meanings understandable by human, then a final step of punctual per-cluster hand-labeling is required. State-of-the-art methods \cite{caron_deep_2018} dealing with complex real images mix trainable feature extractors with standard clustering methods such as a Gaussian Mixture Model (GMM) \cite{moon_expectation_maximization_1996}. 
	\item \textbf{Pre-training:} Some relevant generic visual feature extractors can be obtained by performing a preliminary pre-training of the CNN model on unlabeled or labeled data coming from the target application domain \cite{hadsell_learning_2009} or even from a different one \cite{godard_digging_2018}.
\end{itemize}

We also note that in order to apprehend from end-to-end the learned representations, it is possible to identify training overfitting \cite{houle_local_2017} of deep learning models without validation test supervision. Furthermore, some learning approaches can estimate the prior of a noisy labeled training dataset \cite{jain_estimating_2016} or the model uncertainty \cite{gal_uncertainty_2016}, \cite{kendall_bayesian_2015}.

Now that some considered analytical and learning methods have been treated separately, the next section shows the complementarity between these two different types of approaches through several Self-Supervised Learning (SSL) systems developed in the context the perception of the autonomous driving vehicle.

\section{SSL Autonomous Driving Applications}
\label{section:ssl}

In the context of autonomous driving applications, we can organize the Self-Supervised Learning (SSL) perception techniques in two main categories:
\begin{itemize}
	\item High-level scene understanding:
	\begin{itemize}
		\item road segmentation in order to discriminate the traversable path from obstacles to be avoided
		\item dynamic obstacles detection and segmentation
		\item obstacles tracking and motion anticipation predictions
	\end{itemize}
	\item Low-level sensor data analysis, with a particular focus on: Dense depth map estimation, which is a potentially relevant input data information for dealing with the previously enumerated scene understanding challenges.
\end{itemize}

\subsection{Scene understanding}

In order to navigate safely, smoothly, or swiftly when it is required, a self-driving car must perform a path planning adapted to the surrounding environment. The planned trajectories must pass trough traversable areas, while ensuring that surrounding static and dynamic obstacles are avoided. For this purpose, it is necessary to detect and delineate these potential obstacles in advance, but also to anticipate future positions of the mobile ones.

\subsubsection{Traversable area segmentation}

A traversable area can be identified by performing its segmentation over the mapped physical environment. Two different strategies have been successively applied. The former is mainly dedicated to offroad unknown terrain crossing. It entails fully self-supervised training systems (i.e. without hand-labeled data). The latter, that appeared more recently, is dedicated to urban road analysis. The main difference is that the SSL online systems are initialized with a supervised pre-training on hand-labeled data. This preliminary step aims at replacing the lack of landmarks on urban asphalt roads having uniform textures, by prior knowledge.

\textbf{SSL offroad systems:}
a road segmentation is proposed in \cite{lieb_adaptive_2005} by exploiting temporal past information concerning the road appearance on monocular camera images. 
It considers the close observable area on the current monocular camera frame in front of the car as a traversable road. Next, it propagates optical flow on this area from the current frame up to the past captured frames. Then, it can deduce this close area appearance when it was spatially farther in the past. This past appearance of the actual close traversable area is exploited for producing horizontal line templates using the SSD (sum of squared differences) matching measure. It is combined with a hough transform-based horizon detector to define the image horizontal lines of pixels on which to apply the horizontal 1-D template matching. Next, with the assumption that the actual distant traversable area has roughly the same appearance as the actual close area had in the past, the 1D templates are applied over the current frame to segment the distant traversable area.
If the best template matching measure changes abruptly, then it is supposed that the ego-vehicle is going out of the road or that the road appearance has suddenly and drastically changed. 
The approach in \cite{lieb_adaptive_2005} is relevant for providing a long-range road image segmentation using a monocular camera only. 
However, a major issue is the critical assumption considering the close area as always traversable. If the road aspect changes suddenly, then it is impossible with this SSL strategy to correctly segment this novel road region.

%
Another SSL road segmentation approach is proposed in \cite{dahlkamp_self_supervised_2006} dealing with this issue.
Instead of using temporal information with the assumption that the close area is always traversable, and in addition to the monocular camera, a LIDAR sensor is used for detecting the obstacles close to the ego-vehicle.
Projected on the camera images, LIDAR depth points enable to automatically and sparsely label the close traversable area on images pixels. Then, a learning gaussian mixture model (GMM) is trained online to recognize the statistical appearance of these sparse analytically labeled pixels. Next, the trained model is applied on the camera pixels which cannot benefit from the sparse LIDAR points projection, in order to classify them as road pixels or not. 
In this way, the vehicle can anticipate the far obstacles observable in the monocular camera images, but not in the dispersive LIDAR data. This SSL system enabled the \textit{Stanley} self-driving car, presented in Figure \ref{fig:Self_driving_cars}(a), to win the \textit{DARPA Grand Challenge}\footnote{https://www.darpa.mil/about-us/timeline/-grand-challenge-for-autonomous-vehicles} by smoothing the trajectories and increasing the vehicle speed thanks to the anticipation of distant obstacles. This highlighted the interest of combining multiple sensors in a self-driving car. 

\begin{figure*}[h]
\begin{center}
\begin{minipage}[c]{0.49\linewidth}
  \centering
  \centerline{\includegraphics[width=\linewidth]{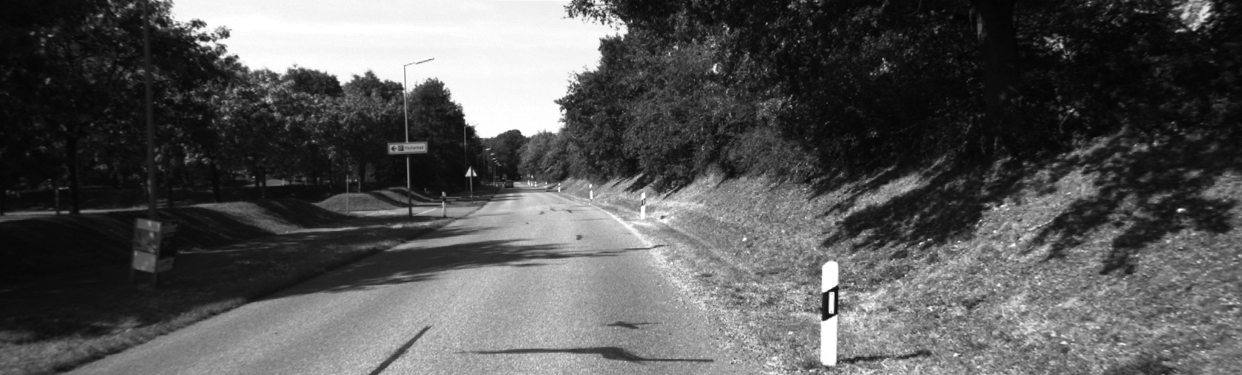}}
  \centerline{(a)}\medskip
\end{minipage}
\begin{minipage}[c]{0.49\linewidth}
  \centering
  \centerline{\includegraphics[width=\linewidth]{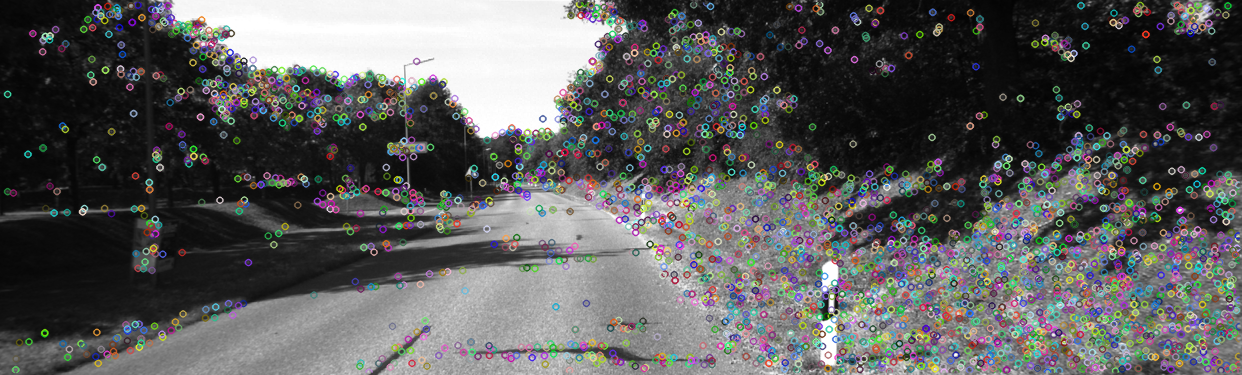}}
  \centerline{(b)}\medskip
\end{minipage}
\caption{Salient features location on urban ego-vehicle environment. (a) is an arbitrary frame, extracted from the KITTI dataset \cite{geiger_are_2012}, illustrating an urban asphalt road with the surrounding environment. (b) shows keypoints detected on the left input image using SIFT detector. Keypoints distribution is dense in the offroad region, and sparse on the asphalt road in the image center.}
\label{fig:sift_urban_road_distr}
\end{center}
%
\end{figure*}

More recently, with the growing interest for deep learning methods, Hadsell et al. \cite{hadsell_learning_2009} propose to use a CNN classifier model instead of the earlier template matching or GMM learning techniques. 
Moreover, an additional paired camera (i.e. stereo-camera) replaces the LIDAR sensor as in \cite{dahlkamp_self_supervised_2006}.
As offroad terrain traversable areas are not always completely flat, a multi-ground plane segmentation is performed in \cite{hadsell_learning_2009}, on the short-range point cloud projection, obtained with the stereo-vision depth map, by using a hough transform plane detector. This technique provides several automatic labels for image patches which are observable in the short-range region. Then, addressing the long-range vision segmentation, the authors firstly train a classifier to predict patches labels automatically estimated within the short-range region. Next, the trained classifier predicts the same labels on the long-range observable image region patches by using a sliding window classification strategy. Concerning the prediction performances, the authors have demonstrated that the online fine tuning of the classifier and the offline pre-taining of its convolutional layers using an unsupervised autoencoder architecture can improve prediction performances. Moreover, an interesting point to note is that instead of using uncertainty or noisy labeled learning techniques, the authors created transition class labels for the boundary image surfaces separating the obstacles from the traversable area.
Finally, from an initial 11-12 meters short range stereo-vision, the developed SSL system is able to extrapolate a long-range vision up to 50-100 meters.  
Nonetheless, in order to estimate the short-range stereo 3D reconstruction, including planar sets of points corresponding to the offroad traversable area, this approach requires the presence of salient visual features in the road regions. This may be impractical for instance on the uniform visual texture of asphalt roads commonly encountered in urban scenarios, as illustrated in Fig. \ref{fig:sift_urban_road_distr}.

\textbf{Pre-trained SSL urban road systems:} Some other online SSL techniques deal with this issue by exploiting a classifer pre-trained offline on hand-labeled data \cite{zhou_road_2010}, \cite{roncancio_traversability_2014}. 

The automatic labeling step previously performed with analytical methods is replaced in \cite{zhou_road_2010} by an SVM classifier pre-trained offline using a human annotated dataset. In this way, this approach is also compatible with uniform asphalt road surfaces. However, compared to the previously presented SSL offroad approaches, it requires hand-labeled data.

A hybrid path segmentation technique is proposed in \cite{roncancio_traversability_2014}. It combines a 3D traversability cost map obtained by stereo-vision, and an SVM classifier pre-trained offline over a human annotated dataset. Six different ground surfaces are considered to train the classifier: asphalt, big gravel, small gravel, soil, grass, bushes and stones. The strategy is as follows. SVM predictions refine online the cost map concerning the flat regions. In turn, the 3D traversability cost map obtained without supervision is exploited to update online some mis-classifications of the pre-trained classifier.

To sum up these road segmentation SSL approaches, we can notice that while the sensor data and the analytical and learning models are different, the online process remains essentially the same. The first stage always consists in generating automatic labels by using additional temporal \cite{lieb_adaptive_2005}, sensor \cite{dahlkamp_self_supervised_2006}, \cite{hadsell_learning_2009}, or prior knowledge information \cite{zhou_road_2010}, \cite{roncancio_traversability_2014}. Then, a second stage trains or updates online a classifier, such that it can be used to provide a long-range or refine road segmentation. 
%
%
%
Overall, while the short-range visions based on depth sensors aims at ensuring the reliable detection of close obstacles, using such SSL vision techniques in static environments directly enables to anticipate the path planning evolution at a long range. Consequently, it is possible to increase the maximum speed velocity of the self-driving car \cite{dahlkamp_self_supervised_2006}, while preserving smooth trajectories \cite{hadsell_learning_2009}.

Now that we have presented some SSL techniques dealing with limited depth sensors in static environments, we focus on dynamic obstacles, as they represent the other potential road users interacting with the ego-vehicle in the shared surrounding environment.

\subsubsection{Dynamic obstacles analysis}

This section starts by presenting an SSL approach \cite{guizilini_online_2013} based on a binary per-pixel segmentation of dynamic obstacles. Then, we introduce its extension \cite{bewley_online_2014} for dynamic obstacles instance segmentation, such that the different road users can be separated.

\textbf{SSL for dynamic obstacles pixel-wise segmentation:} a pixel-level binary segmentation of dynamic obstacles is proposed in \cite{guizilini_online_2013}, using temporal image sequences captured with a monocular camera installed on a mobile urban vehicle.  
The approach firstly separates sparse dynamic keypoints features from the static ones, by applying a RANSAC technique over the optical flow between consecutive frames. Then, the automatically produced per-pixel dynamic labels are transferred as input of a learning Gaussian Process (GP) model. Next, the trained model extrapolates this knowledge to label as dynamic the pixels of the same visual properties instead of the ones previously automatically identified as dynamic. The whole process is achieved during an online procedure.
The system is evaluated on a hand-labeled dataset.
This SSL strategy has the advantage to provide the background subtraction from a moving camera, while extrapolating a dense per-pixel segmentation of the dynamic obstacles from sparse detected keypoints.
However, this technique cannot provide per-obstacles analysis as it merely predicts a binary mask of pixels corresponding to dynamic obstacles.

The technique in \cite{bewley_online_2014} extends the previous approach for SSL multi-instance segmentation by using temporal image sequences captured with a monocular camera installed on a mobile urban vehicle.
The authors apply, over the mobile keypoints detected by \cite{guizilini_online_2013}, a clustering method using the tracked keypoints information such as their spatial location and motion pattern features.
The multi-instance segmentation of dynamic obstacles is evaluated on a hand-labeled video sequence of the KITTI dataset \cite{geiger_are_2012}.

Overall, the authors state that some issues shared with analytical methods persist in their approach. If the dynamic obstacles shadows are projected on the background, then the latter are considered as dynamic as well. Moreover, the segmentation of distant dynamic obstacles can be missed if the corresponding keypoints variations are considered as noise due to the difficulty to detect the corresponding slight optical flow variations. Furthermore, if a dynamic obstacle, either large or close to the sensor, represents the majority of the image keypoints, then this given obstacle is likely to be treated as the static background scene. 

Nonetheless, it is important to bear in mind that these approaches present state-of-the-art competitive performances for dynamic obstacles detection and segmentation without training or pre-training on annotated data. In addition, the method in \cite{bewley_online_2014} provides interesting tools to analyze on the move the dynamic obstacles, for example to separately track them and learn to predict their intention. 

The next focus is on SSL techniques designed for object tracking and temporal predictions in urban road scene evolution, including dynamic obstacles.

\subsubsection{Temporal tracking predictions}

In order to deal with object appearance changes, a competitive SSL tracking technique \cite{kala_tld_2012} proposes an online adaptive strategy combining tracking, learning, and object detector real-time modules. However, in the context of autonomous driving, it may be often necessary to simultaneously track, and even anticipate the trajectories of several surrounding road users. Moreover, being able to consider the interactions between each road user requires some complex motion pattern analysis. 

It turns out that some SSL approaches propose to deal with this challenge by focusing the prediction effort on the entire scene in a unified way, rather than on every obstacles independently.
%
The SSL \textit{deep tracking} system \cite{ondruska_deep_2016}\footnote{Such an approach could be categorized as unsupervised. However, we make the choice in this article to consider that exploiting during the training an additional future temporal information, not available during the prediction step, is a type of self-supervision.} learns to predict the future state of a 2D LIDAR occupancy grid.
This is achieved by training an RNN on the latent space of a CNN autoencoder (AE) which is applied on the input occupancy grid considered as an image. Each cell of the grid is represented by a pixel, which can be color-coded as occluded, void, or as an obstacle surface. Consequently, the model can be trained from end-to-end by learning to predict the next occupancy grid states using the past and current grid states. Solely the prediction output error of non occluded cells is backpropagated during the training.
By definition, this system can perform a self-evaluation by computing a per-pixel photometric error between the predicted occupancy grid and the real future observed occupancy grid at the same temporal instant.
This technique has the advantage of being compatible with complex motion patterns compared to Bayesian and Kalman tracking techniques. In addition, the training process enables to predict the obstacles trajectories even during occlusions. The major interest of \textit{deep tracking} is that, as the model learns to predict a complete scene, it naturally considers interactions between each dynamic obstacle present in the scene.
In \cite{dequaire_deep_2017}, the \textit{deep tracking} model is extended for a real mobile LIDAR sensor by adding a spatial transformer module in order to take into consideration the displacements of the ego-vehicle with respect to its environment during objects tracking.

In turn, these tracking approaches provide the tools to collect motion pattern information of surrounding dynamic obstacles, such that this information may help to classify obstacles depending on their dynamic properties \cite{fathollahi_autonomous_2016}.

\subsection{Low-level sensor data analysis}

This section addresses the sensor data analysis for low-level information estimation in the context of autonomous driving. Compared to the previous methods, the attention has mainly focused recently on SSL depth map estimation from monocular or stereo cameras.

\subsubsection{SSL Depth map estimation}

The self-supervised depth map estimation approach presented in \cite{garg_unsupervised_2016} predicts a depth map from a monocular camera without relying on annotated depth maps.
The pose transformation between both left and right cameras is known. The SSL strategy is as follows. First, the left camera frame is provided as input to a CNN model trained from scratch to predict, the corresponding depth map. Second, an inverse warping is performed by combining the predicted left depth map with the right camera frame in order to output a synthesized frame similar to the input left frame. In this way, an SSL photometric reconstruction error can be computed as output of the decoder part. Next, this per-pixel error is directly used to train the encoder weights using Stochastic Gradient Descent (SGD) optimization technique.
While not requiring pre-training, nor annotated ground-truth depths, this approach presents prediction performances comparable with the state-of-the-art fully supervised monocular techniques.
However, the ground truth pose transformation, related to the inter-view displacement between both cameras, is required. 

Following a similar idea, another technique is proposed in \cite{zhong_self_supervised_2017}. It is trained to reconstruct, from a given frame, the second frame taken from a different point of view. It generates a depth map using a stereo camera during the training step, but also during the prediction step. This makes the approach more robust, such that it becomes competitive with standard stereo matching techniques. 
Moreover, the constraint of preserving two cameras and the pose transformation ground truth for predictions, enables in counterpart to perform online learning. This may be interesting for dealing with novel ego-vehicle environments unseen during the training.

To overcome the necessity of the pose transformation ground-truth, Zhou et al. \cite{zhou_unsupervised_2017} propose to predict, from a temporal sequence of frames, the depth map with a learning model, and the successive camera pose transformations with another learning model. Both models are trained together from end-to-end for making the novel view synthesis of the next frame. 
However, the pose transformation estimation implies that the predicted depth map is defined up to a scale factor. 

A more modular technique \cite{godard_digging_2018} exploits either temporal monocular sequences of frames as in \cite{zhou_unsupervised_2017}, the paired frames of a stereo camera as in \cite{zhong_self_supervised_2017}, or to jointly exploit both temporal and stereo information. This framework also deals with the false depth estimation of moving obstacles by ignoring, during training, the pixels not varying between two consecutive temporal frames. It also deals with occluded pixels when the captured point of view changes by using a minimum re-projection loss.

To summarize, low-level analysis techniques for depth map estimation have demonstrated that SSL strategies without using ground truth labels can bring state-of-the-art solutions competitive with fully supervised techniques.

\begin{figure*}[ht]
%
\begin{center}
\begin{minipage}[c]{0.25\linewidth}
  \centering
  \centerline{\includegraphics[width=\linewidth]{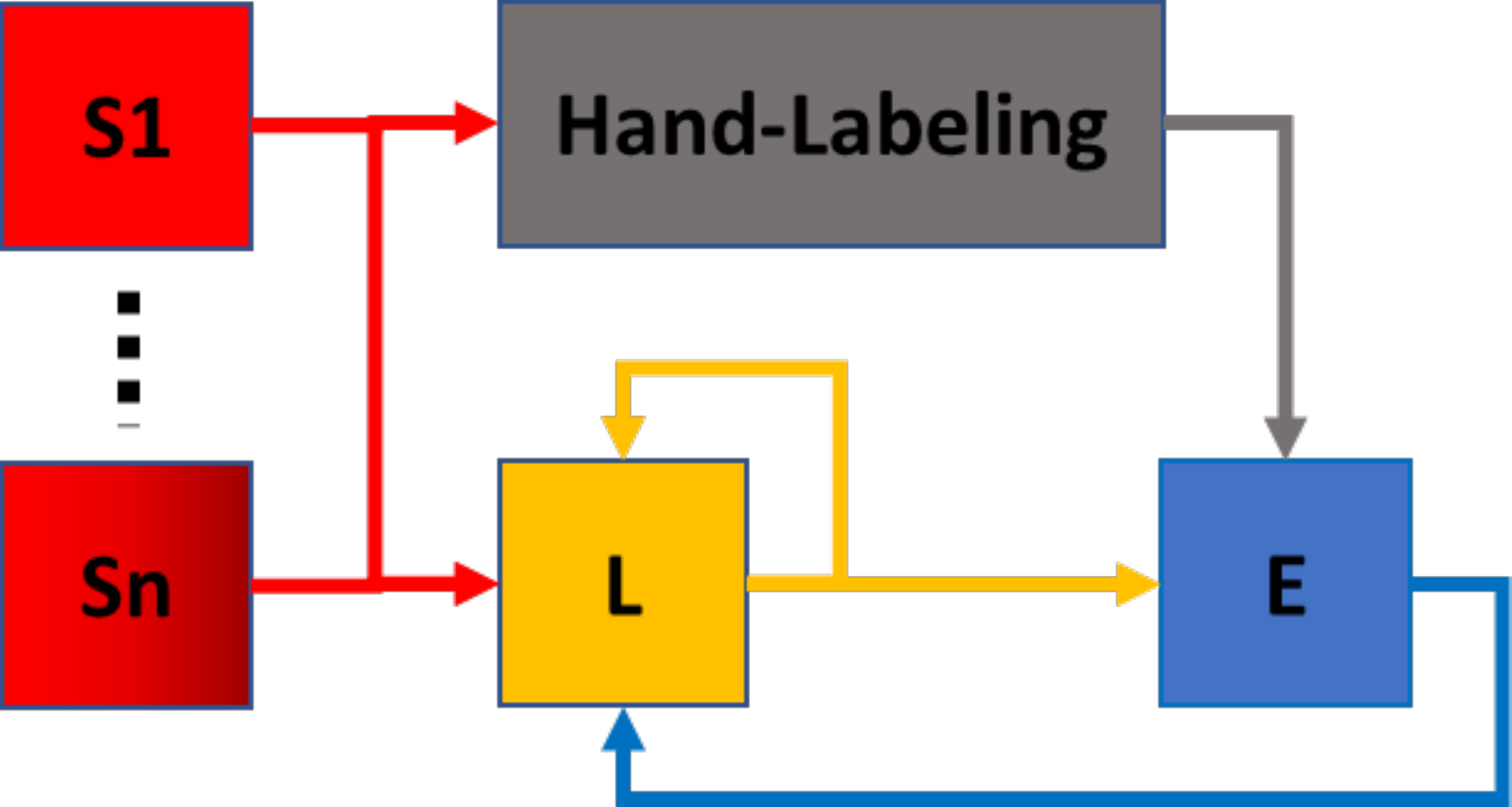}}
\end{minipage}
\begin{minipage}[c]{0.06\linewidth}
  \centering
  \centerline{ }\medskip
\end{minipage}
\begin{minipage}[c]{0.25\linewidth}
  \centering
  \centerline{\includegraphics[width=\linewidth]{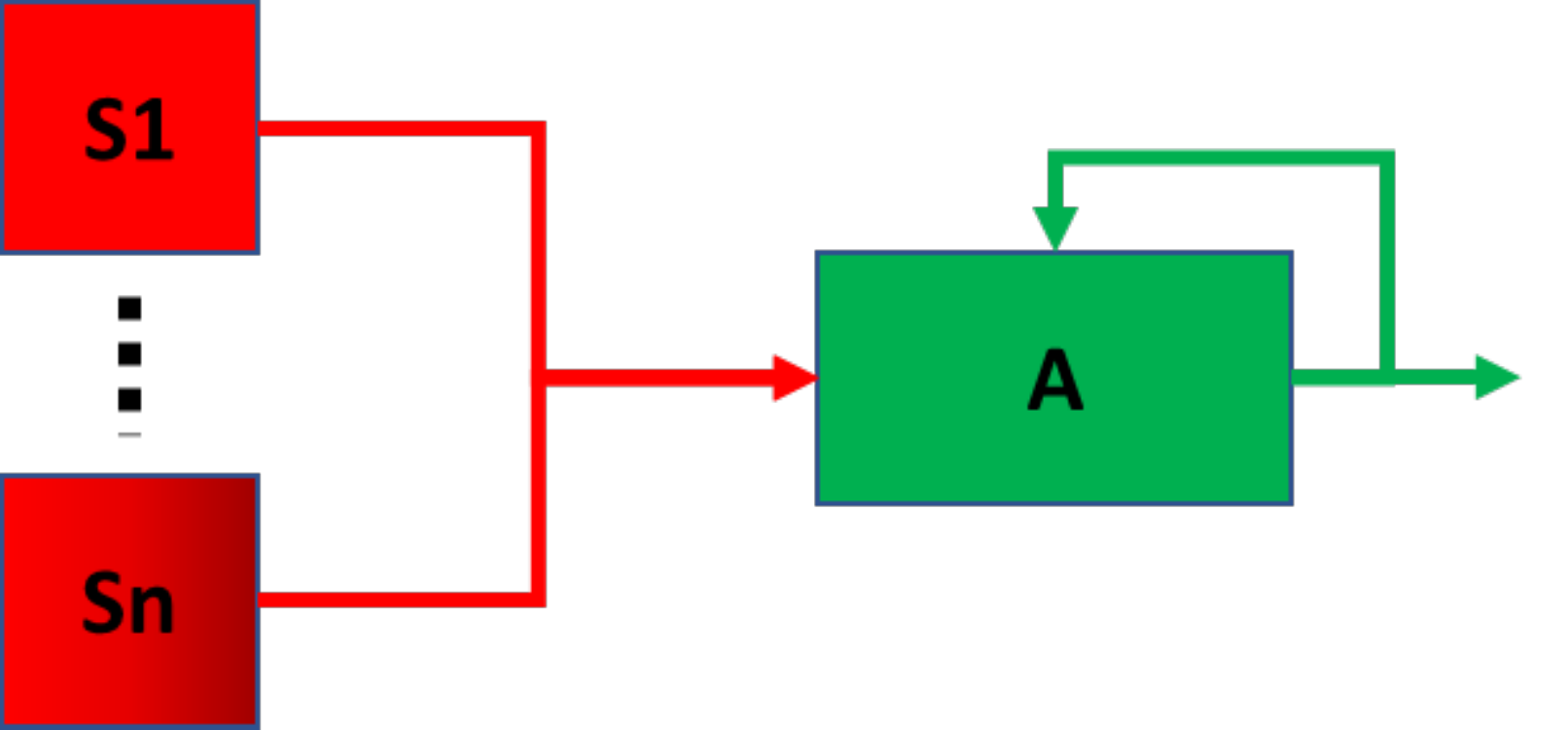}}
\end{minipage}
\begin{minipage}[c]{0.06\linewidth}
  \centering
  \centerline{ }\medskip
\end{minipage}
\begin{minipage}[c]{0.34\linewidth}
  \centering
  \centerline{\includegraphics[width=\linewidth]{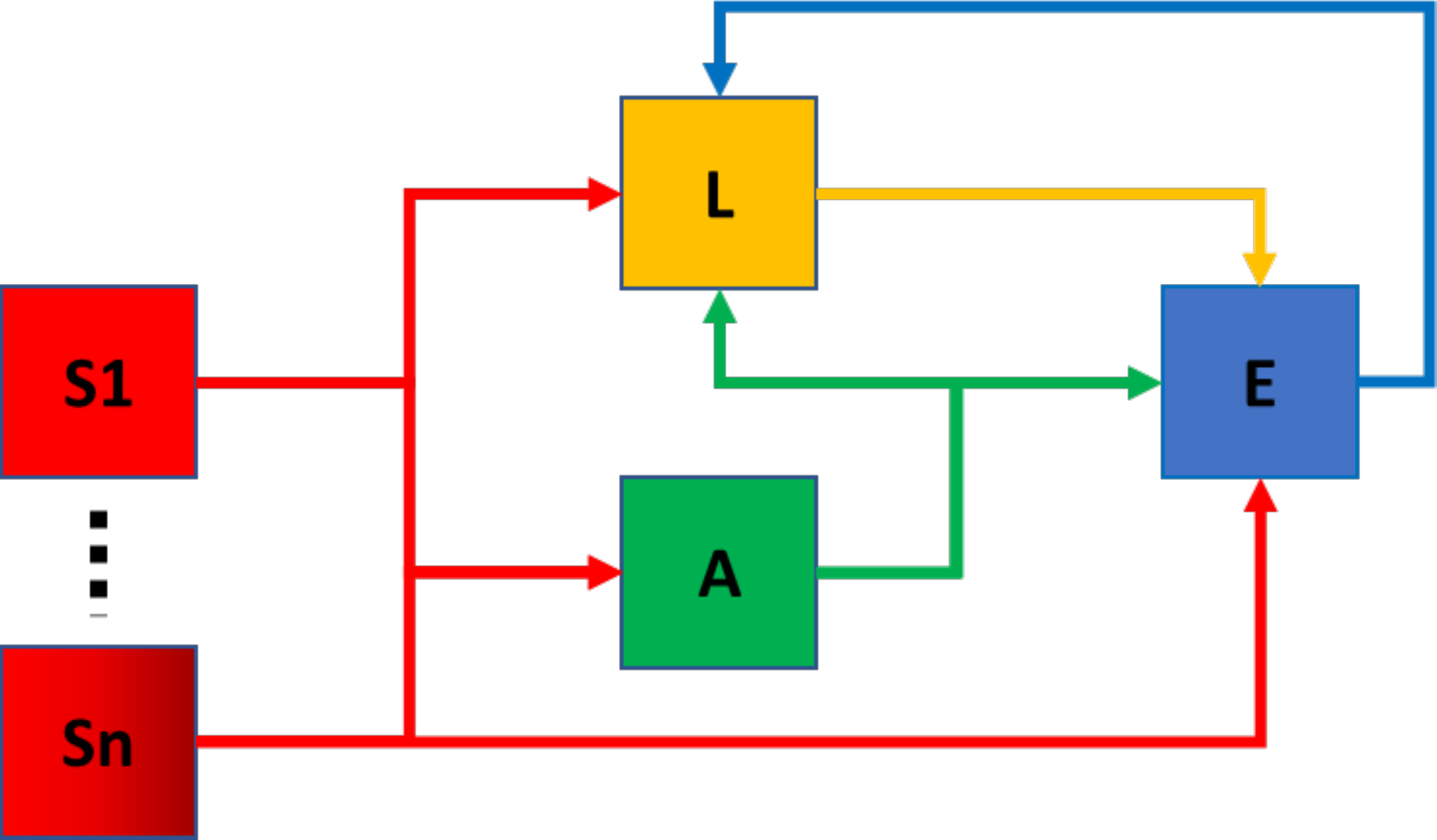}}
\end{minipage}
\begin{minipage}[c]{0.32\linewidth}
  \centering
  \centerline{(a) Supervised learning}\medskip
\end{minipage}
\begin{minipage}[c]{0.32\linewidth}
  \centering
  \centerline{(b) Analytical}\medskip
\end{minipage}
\begin{minipage}[c]{0.32\linewidth}
  \centering
  \centerline{(c) SSL}\medskip
\end{minipage}
\caption{Function diagrams showing the common connections depending on the strategy. Functional blocks represent single monocular camera frame $S_1$, additional sensor data (e.g. temporal frame sequence, stereo-camera, or lidar data) $S_n$, a Learning model $L$, an Analytical method $A$, and Evaluation method $E$.}
\label{fig:SSL_generic_methodology}
\end{center}
\end{figure*}

Overall, the SSL techniques presented in this section support the following conclusion: By exploiting the complementarity between analytical and learning methods, it is possible to address several autonomous driving perception tasks, without necessarily requiring an annotated dataset. Presented methodologies are summarized in Fig. \ref{fig:SSL_generic_methodology} along with Table \ref{tab:Methods_connections}.

\begin{table*}[h]
\centering
\caption{Functional block connections of presented SSL methodologies depending on the application. Experimental datasets exploited and relative prediction performances are reported whenever available. *refers to supervised methods.}
\resizebox{17.75cm}{!}{
\begin{tabular}{c|ccccccccc|c|c}
\toprule
SSL Methodologies &$S_1 \rightarrow L$&$S_n \rightarrow L$&$S_1 \rightarrow A$&$S_n \rightarrow A$&$S_n \rightarrow E$&$A \rightarrow L$&$L \rightarrow E$&$A \rightarrow E$&$E \rightarrow L$& datasets & performances \\
\midrule
\midrule
(Off)road segmentation & & & & & & & & & & & \\
\cite{lieb_adaptive_2005}, \cite{dahlkamp_self_supervised_2006}, \cite{hadsell_learning_2009}, \cite{zhou_road_2010}, \cite{roncancio_traversability_2014} & $\surd$ & & $\surd$ & $\surd$ & & $\surd$ &$\surd$ & $\surd$ & $\surd$ & - & - \\
\midrule
Dynamic obstacles & & & & & & & & & & KITTI \cite{geiger_are_2012} & \\
analysis \cite{guizilini_online_2013}, \cite{bewley_online_2014} & $\surd$ & & $\surd$ & $\surd$ & & $\surd$ & & & & Sidney \cite{guizilini_online_2013} & - \\
\midrule
Temporal tracking & & & & & & & & & &Oxford Robotcar& \\
predictions \cite{ondruska_deep_2016}, \cite{dequaire_deep_2017} & $\surd$ & $\surd$ & & & $\surd$ & & $\surd$ & & $\surd$ & dataset \cite{RobotCarDatasetIJRR} & - \\
\midrule
Depth map estimation & & & & & & & & & & KITTI & \cite{fu2018deep}*$>$\cite{godard_digging_2018}$>$\cite{garg_unsupervised_2016}$>$ \\
\cite{garg_unsupervised_2016}, \cite{zhou_unsupervised_2017}, \cite{zhong_self_supervised_2017}$^1$, \cite{godard_digging_2018}$^1$ & $\surd$ & $\surd^1$ & & & $\surd$ & & $\surd$ & & $\surd$ & Make3D \cite{saxena2008make3d} &\cite{zhou_unsupervised_2017}$>$\cite{eigen2014depth}* \\
\bottomrule
\end{tabular}}
\label{tab:Methods_connections}
\end{table*}

\section{Limitations and future challenges}
\label{section:limit}

In the context of autonomous driving, some limitations remain in the presented SSL perception systems and open future research perspectives.

\textit{Catastrophic forgetting:} During the online learning procedure, the trainable weights of the model may require unnecessary repetitive updates for detecting a given pattern throughout the environment exploration. In fact, when a learning model is continuously specialized for dealing with the latest data, the likelihood increases that the model simultaneously forget the potentially relevant formerly learned patterns. It turns out that it is possible to deal with this \textit{catastrophic forgetting} issue when using neural networks \cite{kirkpatrick_overcoming_2017}. For future research directions, it may be interesting to combine such incremental learning techniques with the presented SSL frameworks.

Concerning the scene depth map estimation solely based on temporal analysis: 
\begin{itemize}
	\item the presence of dynamic obstacles in the scene during the learning stage can result in poor estimates of the observed scene. As discussed in \cite{guizilini_online_2013}, further research on SSL for potentially dynamic obstacles delineations on the sensor data may help to deal with this issue.
	\item the current state-of-the-art techniques cannot estimate the real depth map without requiring a supervised scaling factor. The latter is generally obtained by estimating the real metric values of the pose transformation between two consecutive camera viewpoints. As proposed in the supervised detector \textit{Deep MANTA} \cite{Chabot_2017_CVPR}, it may be interesting to recover this scale factor by using some template matching techniques on the observable objects of the scene.
\end{itemize}

Concerning the online self-evaluation, some of the presented systems require a baseline reference analytically obtained \cite{hadsell_learning_2009}. However, if we consider that the analytical processes, considered as ground-truth labeling techniques, are likely to generate some noisy labels, it may be interesting to investigate some future research on how to evaluate this prior noise from the learning model viewpoint \cite{jain_estimating_2016}, and how to deal with it \cite{chiaroni_hallucinating_2019}.

\ifCLASSOPTIONcaptionsoff
  \newpage
\fi



\bibliographystyle{IEEEtran}
\bibliography{Ma_bibliotheque}

\begin{thebibliography}{10}
\providecommand{\url}[1]{#1}
\csname url@samestyle\endcsname
\providecommand{\newblock}{\relax}
\providecommand{\bibinfo}[2]{#2}
\providecommand{\BIBentrySTDinterwordspacing}{\spaceskip=0pt\relax}
\providecommand{\BIBentryALTinterwordstretchfactor}{4}
\providecommand{\BIBentryALTinterwordspacing}{\spaceskip=\fontdimen2\font plus
\BIBentryALTinterwordstretchfactor\fontdimen3\font minus
  \fontdimen4\font\relax}
\providecommand{\BIBforeignlanguage}[2]{{%
\expandafter\ifx\csname l@#1\endcsname\relax
\typeout{** WARNING: IEEEtran.bst: No hyphenation pattern has been}%
\typeout{** loaded for the language `#1'. Using the pattern for}%
\typeout{** the default language instead.}%
\else
\language=\csname l@#1\endcsname
\fi
#2}}
\providecommand{\BIBdecl}{\relax}
\BIBdecl

\bibitem{lowe_distinctive_2004}
D.~G. Lowe, ``Distinctive image features from scale-invariant keypoints,''
  \emph{International journal of computer vision}, vol.~60, no.~2, pp. 91--110,
  2004.

\bibitem{karpushin_keypoint_2016}
M.~Karpushin, G.~Valenzise, and F.~Dufaux, ``Keypoint detection in rgbd images
  based on an anisotropic scale space,'' \emph{IEEE Transactions on
  Multimedia}, vol.~18, no.~9, pp. 1762--1771, 2016.

\bibitem{bresson_simultaneous_2017}
G.~Bresson, Z.~Alsayed, L.~Yu, and S.~Glaser, ``Simultaneous localization and
  mapping: {A} survey of current trends in autonomous driving,'' \emph{IEEE
  Transactions on Intelligent Vehicles}, vol.~2, no.~3, pp. 194--220, 2017.

\bibitem{kirillov_panoptic_2018}
A.~Kirillov, K.~He, R.~Girshick, C.~Rother, and P.~Dollár, ``Panoptic
  segmentation,'' \emph{arXiv preprint arXiv:1801.00868}, 2018.

\bibitem{cordts_cityscapes_2016}
M.~Cordts, M.~Omran, S.~Ramos, T.~Rehfeld, M.~Enzweiler, R.~Benenson,
  U.~Franke, S.~Roth, and B.~Schiele, ``The cityscapes dataset for semantic
  urban scene understanding,'' in \emph{Proceedings of the {IEEE} conference on
  computer vision and pattern recognition}, 2016, pp. 3213--3223.

\bibitem{niu_theoretical_2016}
G.~Niu, M.~C. du~Plessis, T.~Sakai, Y.~Ma, and M.~Sugiyama, ``Theoretical
  comparisons of positive-unlabeled learning against positive-negative
  learning,'' in \emph{Advances in {Neural} {Information} {Processing}
  {Systems}}, 2016, pp. 1199--1207.

\bibitem{chiaroni_learning_2018}
F.~Chiaroni, M.-C. Rahal, N.~Hueber, and F.~Dufaux, ``Learning with a
  generative adversarial network from a positive unlabeled dataset for image
  classification,'' in \emph{{IEEE} {International} {Conference} on {Image}
  {Processing}}, 2018.

\bibitem{ma_dimensionality-driven_2018}
X.~Ma, Y.~Wang, M.~E. Houle, S.~Zhou, S.~M. Erfani, S.-T. Xia, S.~Wijewickrema,
  and J.~Bailey, ``Dimensionality-driven learning with noisy labels,''
  \emph{arXiv preprint arXiv:1806.02612}, 2018.

\bibitem{chiaroni_hallucinating_2019}
F.~Chiaroni, M.~C. Rahal, N.~Hueber, and F.~Dufaux, ``Hallucinating a {Cleanly}
  {Labeled} {Augmented} {Dataset} from a {Noisy} {Labeled} {Dataset} {Using}
  {GANs},'' in \emph{{IEEE} {International} {Conference} on {Image}
  {Processing}}, 2019.

\bibitem{houle_local_2017}
M.~E. Houle, ``Local intrinsic dimensionality {I}: an extreme-value-theoretic
  foundation for similarity applications,'' in \emph{International {Conference}
  on {Similarity} {Search} and {Applications}}.\hskip 1em plus 0.5em minus
  0.4em\relax Springer, 2017, pp. 64--79.

\bibitem{jain_estimating_2016}
S.~Jain, M.~White, and P.~Radivojac, ``Estimating the class prior and posterior
  from noisy positives and unlabeled data,'' in \emph{Advances in {Neural}
  {Information} {Processing} {Systems} 29}.\hskip 1em plus 0.5em minus
  0.4em\relax Curran Associates, Inc., 2016, pp. 2693--2701.

\bibitem{gal_uncertainty_2016}
Y.~Gal, ``Uncertainty in {Deep} {Learning},'' Ph.D. dissertation, PhD thesis,
  University of Cambridge, 2016.

\bibitem{dahlkamp_self_supervised_2006}
H.~Dahlkamp, A.~Kaehler, D.~Stavens, S.~Thrun, and G.~R. Bradski,
  ``Self-supervised monocular road detection in desert terrain.'' in
  \emph{Robotics: science and systems}, vol.~38.\hskip 1em plus 0.5em minus
  0.4em\relax Philadelphia, 2006.

\bibitem{jing_self-supervised_2019}
L.~Jing and Y.~Tian, ``Self-supervised {Visual} {Feature} {Learning} with
  {Deep} {Neural} {Networks}: {A} {Survey},'' \emph{arXiv preprint
  arXiv:1902.06162}, 2019.

\bibitem{zhong_self_supervised_2017}
Y.~Zhong, Y.~Dai, and H.~Li, ``Self-supervised learning for stereo matching
  with self-improving ability,'' \emph{arXiv preprint arXiv:1709.00930}, 2017.

\bibitem{garg_unsupervised_2016}
R.~Garg, V.~K. BG, G.~Carneiro, and I.~Reid, ``Unsupervised cnn for single view
  depth estimation: {Geometry} to the rescue,'' in \emph{European {Conference}
  on {Computer} {Vision}}.\hskip 1em plus 0.5em minus 0.4em\relax Springer,
  2016, pp. 740--756.

\bibitem{dequaire_deep_2017}
J.~Dequaire, P.~Ondruska, D.~Rao, D.~Wang, and I.~Posner, ``Deep tracking in
  the wild: {End}-to-end tracking using recurrent neural networks,'' \emph{The
  International Journal of Robotics Research}, p. 0278364917710543, 2017.

\bibitem{bewley_online_2014}
\BIBentryALTinterwordspacing
A.~Bewley, V.~Guizilini, F.~Ramos, and B.~Upcroft, ``Online self-supervised
  multi-instance segmentation of dynamic objects,'' in \emph{Robotics and
  {Automation} ({ICRA}), 2014 {IEEE} {International} {Conference} on}.\hskip
  1em plus 0.5em minus 0.4em\relax IEEE, 2014, pp. 1296--1303. [Online].
  Available: \url{http://ieeexplore.ieee.org/abstract/document/6907020/}
\BIBentrySTDinterwordspacing

\bibitem{hadsell_learning_2009}
R.~Hadsell, P.~Sermanet, J.~Ben, A.~Erkan, M.~Scoffier, K.~Kavukcuoglu,
  U.~Muller, and Y.~LeCun, ``Learning long-range vision for autonomous off-road
  driving,'' \emph{Journal of Field Robotics}, vol.~26, no.~2, pp. 120--144,
  2009.

\bibitem{geiger_are_2012}
A.~Geiger, P.~Lenz, and R.~Urtasun, ``Are we ready for autonomous driving? the
  kitti vision benchmark suite,'' in \emph{2012 {IEEE} {Conference} on
  {Computer} {Vision} and {Pattern} {Recognition}}.\hskip 1em plus 0.5em minus
  0.4em\relax IEEE, 2012, pp. 3354--3361.

\bibitem{guizilini_online_2013}
V.~Guizilini and F.~Ramos, ``Online self-supervised segmentation of dynamic
  objects,'' in \emph{Robotics and {Automation} ({ICRA}), 2013 {IEEE}
  {International} {Conference} on}.\hskip 1em plus 0.5em minus 0.4em\relax
  IEEE, 2013, pp. 4720--4727.

\bibitem{pathak_learning_2017}
D.~Pathak, R.~Girshick, P.~Dollár, T.~Darrell, and B.~Hariharan, ``Learning
  features by watching objects move,'' in \emph{Proceedings of the {IEEE}
  {Conference} on {Computer} {Vision} and {Pattern} {Recognition}}, 2017, pp.
  2701--2710.

\bibitem{ondruska_deep_2016}
P.~Ondruska and I.~Posner, ``Deep tracking: {Seeing} beyond seeing using
  recurrent neural networks,'' \emph{arXiv preprint arXiv:1602.00991}, 2016.

\bibitem{harris1988combined}
C.~G. Harris, M.~Stephens \emph{et~al.}, ``A combined corner and edge
  detector.'' in \emph{Alvey vision conference}, vol.~15, no.~50.\hskip 1em
  plus 0.5em minus 0.4em\relax Citeseer, 1988, pp. 10--5244.

\bibitem{bay2006surf}
H.~Bay, T.~Tuytelaars, and L.~Van~Gool, ``Surf: Speeded up robust features,''
  in \emph{European conference on computer vision}.\hskip 1em plus 0.5em minus
  0.4em\relax Springer, 2006, pp. 404--417.

\bibitem{rublee2011orb}
E.~Rublee, V.~Rabaud, K.~Konolige, and G.~R. Bradski, ``Orb: An efficient
  alternative to sift or surf.'' in \emph{ICCV}, vol.~11, no.~1.\hskip 1em plus
  0.5em minus 0.4em\relax Citeseer, 2011, p.~2.

\bibitem{zhong2009intrinsic}
Y.~Zhong, ``Intrinsic shape signatures: A shape descriptor for 3d object
  recognition,'' in \emph{2009 IEEE 12th International Conference on Computer
  Vision Workshops, ICCV Workshops}.\hskip 1em plus 0.5em minus 0.4em\relax
  IEEE, 2009, pp. 689--696.

\bibitem{farneback_two-frame_2003}
G.~Farnebäck, ``Two-frame motion estimation based on polynomial expansion,''
  \emph{Image analysis}, pp. 363--370, 2003.

\bibitem{lucas1981iterative}
B.~D. Lucas, T.~Kanade \emph{et~al.}, ``An iterative image registration
  technique with an application to stereo vision,'' 1981.

\bibitem{menze2015object}
M.~Menze and A.~Geiger, ``Object scene flow for autonomous vehicles,'' in
  \emph{Proceedings of the IEEE Conference on Computer Vision and Pattern
  Recognition}, 2015, pp. 3061--3070.

\bibitem{prazdny1980egomotion}
K.~Prazdny, ``Egomotion and relative depth map from optical flow,''
  \emph{Biological cybernetics}, vol.~36, no.~2, pp. 87--102, 1980.

\bibitem{fischler1981random}
M.~A. Fischler and R.~C. Bolles, ``Random sample consensus: a paradigm for
  model fitting with applications to image analysis and automated
  cartography,'' \emph{Communications of the ACM}, vol.~24, no.~6, pp.
  381--395, 1981.

\bibitem{ballard1981generalizing}
D.~H. Ballard, ``Generalizing the hough transform to detect arbitrary shapes,''
  \emph{Pattern recognition}, vol.~13, no.~2, pp. 111--122, 1981.

\bibitem{scaramuzza2011visual}
D.~Scaramuzza and F.~Fraundorfer, ``Visual odometry [tutorial],'' \emph{IEEE
  robotics \& automation magazine}, vol.~18, no.~4, pp. 80--92, 2011.

\bibitem{davison2007monoslam}
A.~J. Davison, I.~D. Reid, N.~D. Molton, and O.~Stasse, ``Monoslam: Real-time
  single camera slam,'' \emph{IEEE Transactions on Pattern Analysis \& Machine
  Intelligence}, no.~6, pp. 1052--1067, 2007.

\bibitem{Salas_Moreno_2013_CVPR}
R.~F. Salas-Moreno, R.~A. Newcombe, H.~Strasdat, P.~H. Kelly, and A.~J.
  Davison, ``Slam++: Simultaneous localisation and mapping at the level of
  objects,'' in \emph{The IEEE Conference on Computer Vision and Pattern
  Recognition (CVPR)}, June 2013.

\bibitem{teichmann_multinet:_2016}
\BIBentryALTinterwordspacing
M.~Teichmann, M.~Weber, M.~Zoellner, R.~Cipolla, and R.~Urtasun, ``{MultiNet}:
  {Real}-time {Joint} {Semantic} {Reasoning} for {Autonomous} {Driving},''
  \emph{arXiv preprint arXiv:1612.07695}, 2016. [Online]. Available:
  \url{https://arxiv.org/abs/1612.07695}
\BIBentrySTDinterwordspacing

\bibitem{ren_faster_2015}
\BIBentryALTinterwordspacing
S.~Ren, K.~He, R.~Girshick, and J.~Sun, ``Faster {R}-{CNN}: {Towards} real-time
  object detection with region proposal networks,'' in \emph{Advances in neural
  information processing systems}, 2015, pp. 91--99. [Online]. Available:
  \url{http://papers.nips.cc/paper/5638-faster-r-cnn-towards-real-time-object-detection-with-region-proposal-networks}
\BIBentrySTDinterwordspacing

\bibitem{redmon_you_2016}
J.~Redmon, S.~Divvala, R.~Girshick, and A.~Farhadi, ``You only look once:
  {Unified}, real-time object detection,'' in \emph{Proceedings of the {IEEE}
  conference on computer vision and pattern recognition}, 2016, pp. 779--788.

\bibitem{garcia2018survey}
A.~Garcia-Garcia, S.~Orts-Escolano, S.~Oprea, V.~Villena-Martinez,
  P.~Martinez-Gonzalez, and J.~Garcia-Rodriguez, ``A survey on deep learning
  techniques for image and video semantic segmentation,'' \emph{Applied Soft
  Computing}, vol.~70, pp. 41--65, 2018.

\bibitem{He_2017_ICCV}
K.~He, G.~Gkioxari, P.~Dollar, and R.~Girshick, ``Mask r-cnn,'' in \emph{The
  IEEE International Conference on Computer Vision (ICCV)}, Oct 2017.

\bibitem{Dosovitskiy_2015_ICCV}
A.~Dosovitskiy, P.~Fischer, E.~Ilg, P.~Hausser, C.~Hazirbas, V.~Golkov,
  P.~van~der Smagt, D.~Cremers, and T.~Brox, ``Flownet: Learning optical flow
  with convolutional networks,'' in \emph{The IEEE International Conference on
  Computer Vision (ICCV)}, December 2015.

\bibitem{deep_sup_depth_map_est}
F.~{Liu}, C.~{Shen}, G.~{Lin}, and I.~{Reid}, ``Learning depth from single
  monocular images using deep convolutional neural fields,'' \emph{IEEE
  Transactions on Pattern Analysis and Machine Intelligence}, vol.~38, no.~10,
  pp. 2024--2039, Oct 2016.

\bibitem{milan_online_2017}
A.~Milan, S.~H. Rezatofighi, A.~Dick, I.~Reid, and K.~Schindler, ``Online
  multi-target tracking using recurrent neural networks,'' in
  \emph{Thirty-{First} {AAAI} {Conference} on {Artificial} {Intelligence}},
  2017.

\bibitem{caron_deep_2018}
M.~Caron, P.~Bojanowski, A.~Joulin, and M.~Douze, ``Deep clustering for
  unsupervised learning of visual features,'' in \emph{Proceedings of the
  {European} {Conference} on {Computer} {Vision} ({ECCV})}, 2018, pp. 132--149.

\bibitem{moon_expectation_maximization_1996}
T.~K. Moon, ``The expectation-maximization algorithm,'' \emph{IEEE Signal
  processing magazine}, vol.~13, no.~6, pp. 47--60, 1996.

\bibitem{godard_digging_2018}
C.~Godard, O.~Mac~Aodha, and G.~Brostow, ``Digging into self-supervised
  monocular depth estimation,'' \emph{arXiv preprint arXiv:1806.01260}, 2018.

\bibitem{kendall_bayesian_2015}
\BIBentryALTinterwordspacing
A.~Kendall, V.~Badrinarayanan, and R.~Cipolla, ``Bayesian segnet: {Model}
  uncertainty in deep convolutional encoder-decoder architectures for scene
  understanding,'' \emph{arXiv preprint arXiv:1511.02680}, 2015. [Online].
  Available: \url{https://arxiv.org/abs/1511.02680}
\BIBentrySTDinterwordspacing

\bibitem{lieb_adaptive_2005}
D.~Lieb, A.~Lookingbill, and S.~Thrun, ``Adaptive {Road} {Following} using
  {Self}-{Supervised} {Learning} and {Reverse} {Optical} {Flow}.'' in
  \emph{Robotics: science and systems}, 2005, pp. 273--280.

\bibitem{zhou_road_2010}
S.~Zhou, J.~Gong, G.~Xiong, H.~Chen, and K.~Iagnemma, ``Road detection using
  support vector machine based on online learning and evaluation,'' in
  \emph{2010 {IEEE} {Intelligent} {Vehicles} {Symposium}}.\hskip 1em plus 0.5em
  minus 0.4em\relax IEEE, 2010, pp. 256--261.

\bibitem{roncancio_traversability_2014}
H.~Roncancio, M.~Becker, A.~Broggi, and S.~Cattani, ``Traversability analysis
  using terrain mapping and online-trained terrain type classifier,'' in
  \emph{2014 {IEEE} {Intelligent} {Vehicles} {Symposium} {Proceedings}}.\hskip
  1em plus 0.5em minus 0.4em\relax IEEE, 2014, pp. 1239--1244.

\bibitem{kala_tld_2012}
Z.~{Kalal}, K.~{Mikolajczyk}, and J.~{Matas}, ``Tracking-learning-detection,''
  \emph{IEEE Transactions on Pattern Analysis and Machine Intelligence},
  vol.~34, no.~7, pp. 1409--1422, July 2012.

\bibitem{fathollahi_autonomous_2016}
M.~Fathollahi and R.~Kasturi, ``Autonomous driving challenge: {To} {Infer} the
  property of a dynamic object based on its motion pattern using recurrent
  neural network,'' \emph{arXiv preprint arXiv:1609.00361}, 2016.

\bibitem{zhou_unsupervised_2017}
T.~Zhou, M.~Brown, N.~Snavely, and D.~G. Lowe, ``Unsupervised learning of depth
  and ego-motion from video,'' in \emph{Proceedings of the {IEEE} {Conference}
  on {Computer} {Vision} and {Pattern} {Recognition}}, 2017, pp. 1851--1858.

\bibitem{RobotCarDatasetIJRR}
\BIBentryALTinterwordspacing
W.~Maddern, G.~Pascoe, C.~Linegar, and P.~Newman, ``{1 Year, 1000km: The Oxford
  RobotCar Dataset},'' \emph{The International Journal of Robotics Research
  (IJRR)}, vol.~36, no.~1, pp. 3--15, 2017. [Online]. Available:
  \url{http://dx.doi.org/10.1177/0278364916679498}
\BIBentrySTDinterwordspacing

\bibitem{fu2018deep}
H.~Fu, M.~Gong, C.~Wang, K.~Batmanghelich, and D.~Tao, ``Deep ordinal
  regression network for monocular depth estimation,'' in \emph{Proceedings of
  the IEEE Conference on Computer Vision and Pattern Recognition}, 2018, pp.
  2002--2011.

\bibitem{saxena2008make3d}
A.~Saxena, M.~Sun, and A.~Y. Ng, ``Make3d: Learning 3d scene structure from a
  single still image,'' \emph{IEEE transactions on pattern analysis and machine
  intelligence}, vol.~31, no.~5, pp. 824--840, 2008.

\bibitem{eigen2014depth}
D.~Eigen, C.~Puhrsch, and R.~Fergus, ``Depth map prediction from a single image
  using a multi-scale deep network,'' in \emph{Advances in neural information
  processing systems}, 2014, pp. 2366--2374.

\bibitem{kirkpatrick_overcoming_2017}
J.~Kirkpatrick, R.~Pascanu, N.~Rabinowitz, J.~Veness, G.~Desjardins, A.~A.
  Rusu, K.~Milan, J.~Quan, T.~Ramalho, and A.~Grabska-Barwinska, ``Overcoming
  catastrophic forgetting in neural networks,'' \emph{Proceedings of the
  national academy of sciences}, p. 201611835, 2017.

\bibitem{Chabot_2017_CVPR}
F.~Chabot, M.~Chaouch, J.~Rabarisoa, C.~Teuliere, and T.~Chateau, ``Deep manta:
  A coarse-to-fine many-task network for joint 2d and 3d vehicle analysis from
  monocular image,'' in \emph{The IEEE Conference on Computer Vision and
  Pattern Recognition (CVPR)}, July 2017.

\end{thebibliography}
\end{document}